\documentclass[review]{elsarticle}

\usepackage{lineno,hyperref}
\usepackage{mathrsfs}

\usepackage{algpseudocode}  
\usepackage{amsmath}  
\usepackage{amsfonts}

\usepackage{pgfplots}
\usepackage{caption}
\usepackage{subcaption}
\usepackage{microtype}
\usepackage{multirow}
\usepackage{colortbl}
\usepackage{amssymb}
\usepackage[inkscapelatex=false]{svg}
\usepackage[linesnumbered,ruled,boxed,norelsize,vlined,commentsnumbered]{algorithm2e}

\usepackage{amsmath}
\usepackage{amsthm}
\usepackage{adjustbox}
\usepackage{booktabs}

\newcommand{\tbl}[1]{\textcolor{black}{#1}}
\newcommand{\td}[1]{\textcolor{black}{#1}}
\newcommand{\tgray}[1]{\textcolor{gray}{#1}}

\journal{Journal of \LaTeX\ Templates}
\biboptions{authoryear}

\begin{document}

\begin{frontmatter}

\title{Adaptive Patch Contrast for Weakly Supervised Semantic Segmentation}
		
		\author[firstaddress,secondaddress]{Wangyu Wu}\ead{wangyu.wu22@student.xjtlu.edu.cn}
            \author[thirdaddress]{Tianhong Dai}\ead{tianhongdai914@gmail.com}
            \author[fifthaddress]{Zhenhong Chen}\ead{zcheh@microsoft.com}
            \author[secondaddress]{Xiaowei Huang}\ead{xiaowei.huang@liverpool.ac.uk}
		\author[firstaddress]{Jimin Xiao}\ead{jimin.xiao@xjtlu.edu.cn}
  
            \author[firstaddress]{Fei Ma\corref{mycorrespondingauthor}}
		\cortext[mycorrespondingauthor]{Corresponding authors} \ead{fei.ma@xjtlu.edu.cn}
  \author[fourthaddress]{Renrong Ouyang\corref{mycorrespondingauthor}}\ead{szjsjt_oyrr@163.com}

		\address[firstaddress]{Xi'an Jiaotong-Liverpool University, Suzhou, China}
		\address[secondaddress]{University of Liverpool, Liverpool, UK}
            \address[thirdaddress]{University of Aberdeen, Aberdeen, UK}
            \address[fourthaddress]{Suzhou Construction and Transportation Branch, Jiangsu United Vocational and Technical College, Suzhou, China}
            \address[fifthaddress]{Microsoft, Redmond, USA}

\begin{abstract}
\tbl{Weakly Supervised Semantic Segmentation (WSSS), using only image-level labels, has garnered significant attention due to its cost-effectiveness. Typically, the framework} involves using image-level labels as training data to generate pixel-level pseudo-labels with refinements. Recently, methods based on Vision Transformers (ViT) have demonstrated superior capabilities in generating reliable pseudo-labels, particularly in recognizing complete object regions. However, current ViT-based approaches have some limitations in the use of patch embeddings, being prone to being dominated by certain abnormal patches, as well as many multi-stage methods being time-consuming and lengthy in training, thus lacking efficiency. Therefore, in this paper, we introduce a novel ViT-based WSSS method named \textit{Adaptive Patch Contrast} (APC) that significantly enhances patch embedding learning for improved segmentation effectiveness. APC utilizes an Adaptive-K Pooling (AKP) layer to address the limitations of previous max pooling selection methods. Additionally, we propose a Patch Contrastive Learning (PCL) to enhance patch embeddings, thereby further improving the final results. \tbl{We developed an end-to-end single-stage framework without CAM, which improved training efficiency. Experimental results demonstrate that our method performs exceptionally well on public datasets, outperforming other state-of-the-art WSSS methods with a shorter training time.}
\end{abstract}


\end{frontmatter}


\section{Introduction}\label{sec1}

Semantic segmentation~\cite{yudin2024hierarchical,wang2023novel,zhang2021affinity} is an important task in the field of computer vision~\cite{yin2023semi,jiang2024mfdnet,li2024high-fidelity,guo2024dual-hybrid,yin2024class,yin2024class2}. Weakly Supervised Semantic Segmentation (WSSS) is a continuously evolving approach in this field, which aims to generate pixel-level labels by utilizing weak supervision signals to significantly reduce the cost of annotations. Classic weak supervision signals include image-level labels~\cite{ahn2018learning,jang2023weakly,xu2024mctformer+,zhang2024weakly}, points~\cite{bearman2016s}, scribbles~\cite{zhang2021dynamic}, and bounding boxes~\cite{lee2021bbam}. Among the various weak supervision signals, most recent research has focused on the image-level label, primarily because it is the cheapest and contains the least information. This work also falls within the domain of WSSS, where it exclusively utilizes image-level labels. 

\begin{figure}[t]
\centering
\includegraphics[width=0.9\linewidth]{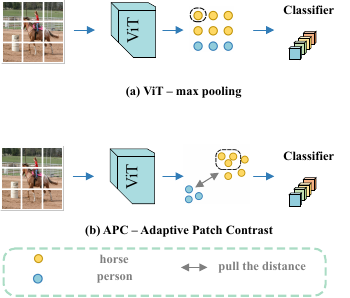}

\caption{In the case of predicting the specific category 'horse': \tbl{(a) Previous ViT-based method~\cite{dosovitskiy2020image} uses the highest-scoring patch. (b) Our APC uses adaptive $K$ patches and patch contrastive learning to enhance patch embeddings for image-level classification.}}
\label{fig:idea}
\vspace{-0.3cm} 
\end{figure}
Prevalent works of WSSS approaches relying on image-level class labels typically generate pseudo labels using class activation maps (CAM)~\cite{zhou2016learning}. However, CAMs have limitations in accurately estimating both the shape and localization of objects belonging to the classes of interest~\cite{chen2022class}. This has prompted researchers to incorporate additional refinements between the initial pseudo labels and the final pseudo labels generation. These refinements often involve multi-stage architectures, as observed in PAMR~\cite{araslanov2020single}, thereby increasing complexity. Notable refinement strategies have been described in IRNet~\cite{ahn2019weakly} and AdvCAM~\cite{lee2021anti}. However, these multi-stage architectures greatly impact computational performance, leading to inefficient training.
In recent years, due to the limitations of CAM, researchers have turned to leverage ViT-based frameworks for WSSS~\cite{rossetti2022max,ru2023token,xu2022multi,ru2022learning}. ViT-PCM~\cite{rossetti2022max} utilizes patch embeddings to infer the probability of pixel-level labels. Methods such as~\cite{ru2023token,xu2022multi} use ViT to replace CNN, thereby enhancing CAM's capability in object recognition. Additionally, AFA~\cite{ru2022learning} proposes an end-to-end architecture based on CAM and suggests acquiring reliable semantic affinity through attention blocks to enhance the initial coarse labels. However, current ViT-based methods utilize global max pooling to select the patch with the highest prediction score, projecting patch-level classification to image-level, which may impact final performance due to misclassification. Moreover, frameworks without CAM are still in multi-stage training, requiring the separate generation of initial pseudo labels before proceeding to the next stage of training, which lacks efficiency




In this work, we empirically observe that existing ViT frameworks without CAM utilize a max pooling layer to connect image embedding patches with softmax, representing the probabilities of different image categories. However, max pooling has limitations as it may be dominated by individual outlier patches, lacking robustness. Therefore, we propose an effective approach called \textit{Adaptive K pooling with patch contrastive learning}, as illustrated in Figure~\ref{fig:idea}, to address the aforementioned issue. Firstly, we propose an Adaptive-K Pooling (AKP) module to overcome the limitations of max pooling by replacing it with a adaptive-k pooling layer. This helps in mitigating the influence of outlier patches and better represents the contribution of multiple different regions of objects to the final prediction. Specifically, in the AKP module, the original max pooling is replaced with a top-K pooling layer, where an adaptive algorithm based on numerical differences selects the optimal K value, addressing the issue of single-point dependence. Furthermore, to effectively combat the issue of over-smoothing, we propose a Patch Contrastive Learning (PCL) module to enhance intra-class compactness and inter-class separability, thereby generating more accurate patch predictions. In the PCL module, we calculate pairwise cosine similarities for patch embeddings generated by ViT, bringing closer the distances in the embedding space for the same category and increasing the distances between different classes, effectively improving the final performance. \tbl{Building on the proposed AKP and PCL modules, we develop an end-to-end, single-stage WSSS framework without CAM, achieving notable improvements in computational efficiency.}

Overall, the contributions of our work can be summarized as follows:


1) \tbl{We propose Adaptive K Pooling (AKP) to address the limitation of segmentation performance being dominated by individual patches. The AKP method mitigates the influence of outlier patches and better represents the contribution of multiple different regions of objects to the final prediction.}

2) We propose a patch-based contrastive learning module, in which we introduce the Patch Contrastive Learning (PCL). By computing the cosine similarity between patch pairs, we aim to increase the distance between patch pairs of different classes and reduce the distance between patch pairs of the same class, thereby enhancing the intra-class compactness and inter-class separability of patch embeddings and further improving the quality of pseudo labels.

3) \tbl{We propose an end-to-end single-stage training framework based on ViT without CAM. This framework effectively enhances the training efficiency of WSSS. The Experiments on the PASCAL VOC 2012~\cite{everingham2010pascal} and MS COCO 2014 datasets~\cite{lin2014microsoft} show that the proposed APC outperforms other state-of-the-art WSSS methods with a shorter training time.}

\tbl{The remainder of this paper is organized into the following sections: Section~\ref{sec2} provides a summary of the related work, while Section~\ref{sec3} presents the proposed method. Section~\ref{sec4} details the experimental dataset, environmental setup, and outcomes. Finally, the conclusion, limitations of our work, and future directions are elaborated in Section~\ref{sec6}.}

\section{Related work}\label{sec2}

\subsection{WSSS with Image-level Labels}
Existing WSSS methods commonly rely on image-level class labels as the cheapest form of supervision. Approaches using image-level class labels have traditionally been based on CAM methods~\cite{zhou2016learning}, employing a standard multi-label classification network. 
The CAMs are derived by applying global average pooling (GAP) to the feature maps of the last layer, followed by concatenation into a weights vector. This vector is then connected to the class prediction through Binary Cross-Entropy (BCE) prediction loss. A common limitation of CAM is its tendency to activate only the most discriminative object regions. To address this limitation, recent studies have proposed various training strategies, including techniques such as erasing~\cite{wei2017object}, online attention accumulation~\cite{jiang2019integral}, and cross-image semantic mining~\cite{sun2020mining}. Researchers in~\cite{chang2020weakly} suggest leveraging auxiliary tasks to regularize the training objective, such as learning visual words. Contrast pixel and prototype representations~\cite{chen2022self,du2022weakly} to promote the comprehensive activation of object regions. To enable the network to capture more object parts, researchers have introduced greater challenges to the classification objective. This has been achieved by modifying either the input data~\cite{singh2017hide,wei2017object,wu2024image,zhang2021complementary,li2018tell} or the feature maps~\cite{lee2019ficklenet,hou2018self,choe2019attention}, employing techniques like dropping out parts of the image or introducing perturbations. Some studies~\cite{chang2020weakly} have made the classification task more difficult by adding finer-grained categories. Additionally, information across multiple images has been used to improve CAM maps~\cite{fan2020cian,sun2020mining}.

\tbl{Typically, the limitation of these methods stems from their dependence on the CAM framework, which focuses primarily on the most discriminative object regions. To address this, we propose a novel WSSS framework that operates without CAM.} We utilize Vision Transformer (ViT) to generate image patch embeddings and then utilize Adaptive K pooling to predict the categories of each patch. Finally, we map the patches to pixels as the segmentation result to address the limitation of CAM activating only the most discriminative object regions.

\subsection{\tbl{Vision Transformers for WSSS}}
Vision Transformer (ViT)\cite{dosovitskiy2020image} has achieved notable success across a range of vision tasks\cite{carion2020end,dosovitskiy2020image}. This success has led to the adoption of ViT in Weakly Supervised Semantic Segmentation (WSSS), where ViT-based methods have begun to emerge as alternatives to traditional CNN-based approaches for generating Class Activation Maps (CAMs)\cite{xu2022multi,ru2022learning}. \tbl{Among these methods are models like MCTformer~\cite{xu2022multi} and AFA~\cite{ru2022learning}, which, despite leveraging the strengths of ViT, still rely on CAMs. They utilize ViT’s multi-head self-attention to capture global contextual information and employ an affinity module to propagate pseudo-masks. However, these approaches continue to face the inherent challenge of over-smoothing in ViT. ViT-PCM~\cite{rossetti2022max} moves away from the reliance on CAMs by employing patch embeddings and max pooling to infer pixel-level label probabilities. This represents a notable advancement, as it is the first framework in WSSS that does not depend on CAMs to generate baseline pseudo-masks. Nonetheless, its reliance on max pooling introduces potential issues, particularly when patches are misclassified. This highlights a critical limitation in existing approaches, where either architectural adjustments or additional modules are required to mitigate the drawbacks of over-smoothing and misclassification.}

In contrast to these methods, our approach employs ViT as the backbone and incorporates adaptive K pooling for the first time. This innovation addresses the limitations associated with max pooling and the potential for misclassification of patches within a framework that does not rely on CAM. Additionally, we introduce the Patch-level Contrastive Learning (PCL) module, which enhances intra-class compactness and inter-class separability of patch embeddings. This further improves the quality of the final labels, providing a more robust and accurate framework for WSSS without relying on CAMs.

\subsection{Single-stage WSSS methods}
Single-stage WSSS methods unify multiple stages such as classification, pseudo-label refinement, and segmentation into a single joint training process, greatly enhancing training efficiency compared to the previous multi-stage frameworks. 1Stage~\cite{araslanov2020single} achieves performance comparable to mainstream multi-stage methods by ensuring local consistency, semantic fidelity, and mask completeness. AFA~\cite{ru2022learning} explores the intrinsic architecture of ViT and extracts reliable semantic affinity from multi-head self-attention for pseudo-label refinement. ToCo~\cite{ru2023token} addresses the observed issue of excessive smoothing in ViT by supervising the final patch tokens with intermediate knowledge. \tbl{These works have pioneered effective single-stage, end-to-end frameworks, offering a more concise and efficient paradigm for WSSS tasks. However, previous single-stage frameworks are fundamentally built upon CAM as a core component, limiting their potential due to CAM's inherent drawback of activating only the most discriminative object regions. In contrast to these single-stage methods, we propose a novel framework that eliminates the reliance on CAM. By employing patch-level predictions directly to pixels, our approach overcomes the limitations of CAM-based methods and achieves state-of-the-art performance in WSSS.
}

\begin{figure*}[h] 
\begin{center}
    \includegraphics[width=1.0\linewidth]{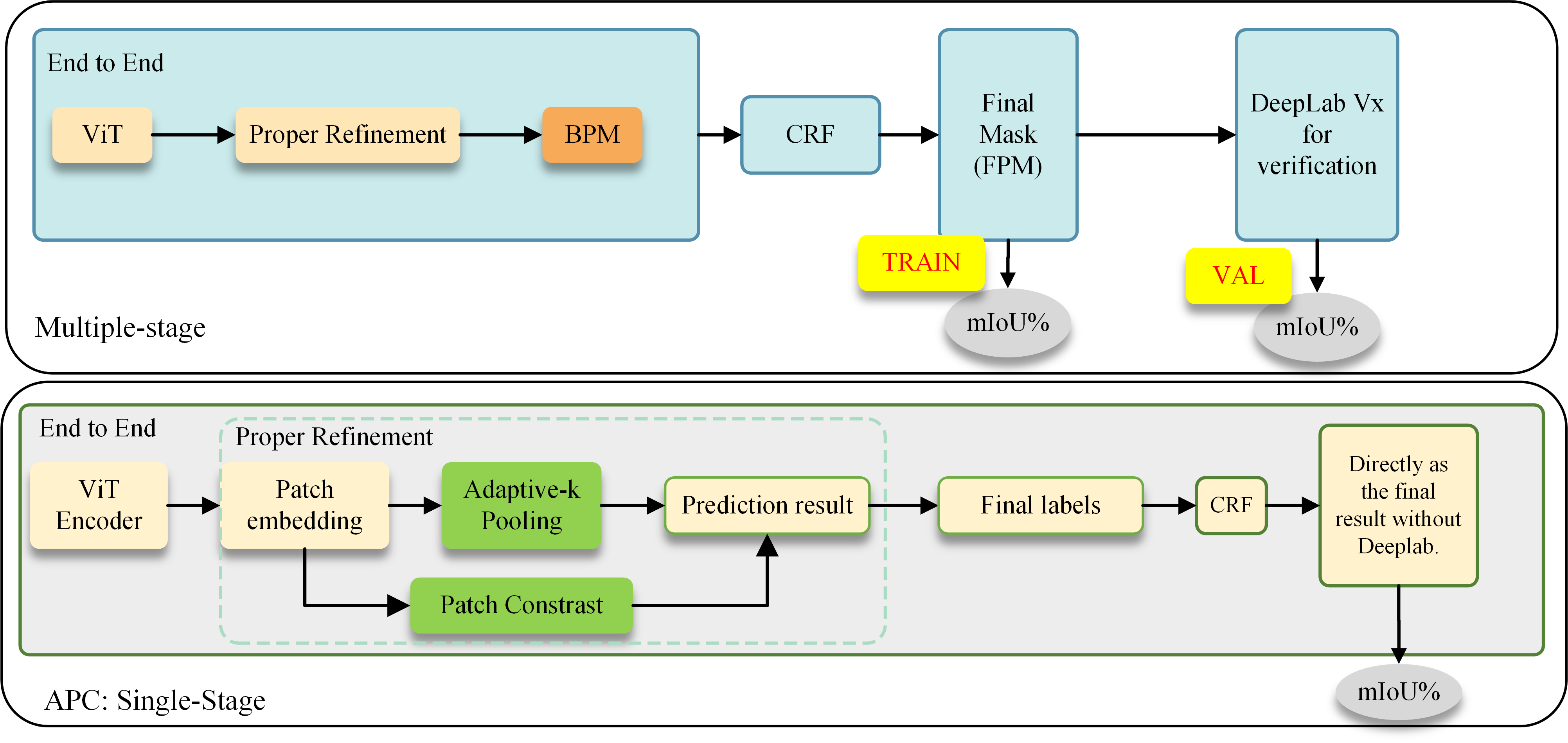}

   \caption{\tbl{This figure compares the basic structure of a multiple-stage WSSS method (light blue) with our APC method (light gray). APC, based on the ViT framework without CAM, allows for a single-stage approach, enabling direct final label acquisition for verification without an additional semantic segmentation model (e.g., DeepLab~\cite{chen2017deeplab}).}}
    \label{fig:framework}
\end{center}
\end{figure*}
\section{Method}\label{sec3}

In this section, we introduce the overall structure and key components of our proposed method. Initially, in Section~\ref{sec:Overview}, we provide an overview of the APC method and compare it with the existing multi-stage approaches. Subsequently, in Section~\ref{sec:model}, we detail the specifics of our model framework, including how to obtain image patch embeddings from images and how to perform segmentation tasks. We have enhanced the existing non-CAM multi-stage framework into a single-stage framework, incorporating a segmentation decoder that utilizes a decoder head to merge multi-level feature maps for prediction, implemented through simple MLP layers. In Section~\ref{sec:top-K_Pooling}, we describe the proposed adaptive K pooling layer, which selects the final prediction patches by choosing different K values based on the ratio of prediction scores, aiming to address incorrect predictions dominated by a single patch. Furthermore, in Section~\ref{sec:PatchContrast}, we discuss the integration of contrastive learning into our current framework. We introduce patch-level contrastive learning, calculating the cosine similarity for pairs of patch embeddings produced by ViT. We define patch contrastive learning (PCL) to enhance patch representations, reduce the distance between patches of the same category, and increase the distance between patches of different categories, thereby further improving prediction accuracy. Finally, in Section~\ref{sec:loss}, we will summarize our APC overall loss.

\begin{figure*}[t] 

\begin{center}
    \includegraphics[width=1.0\linewidth]{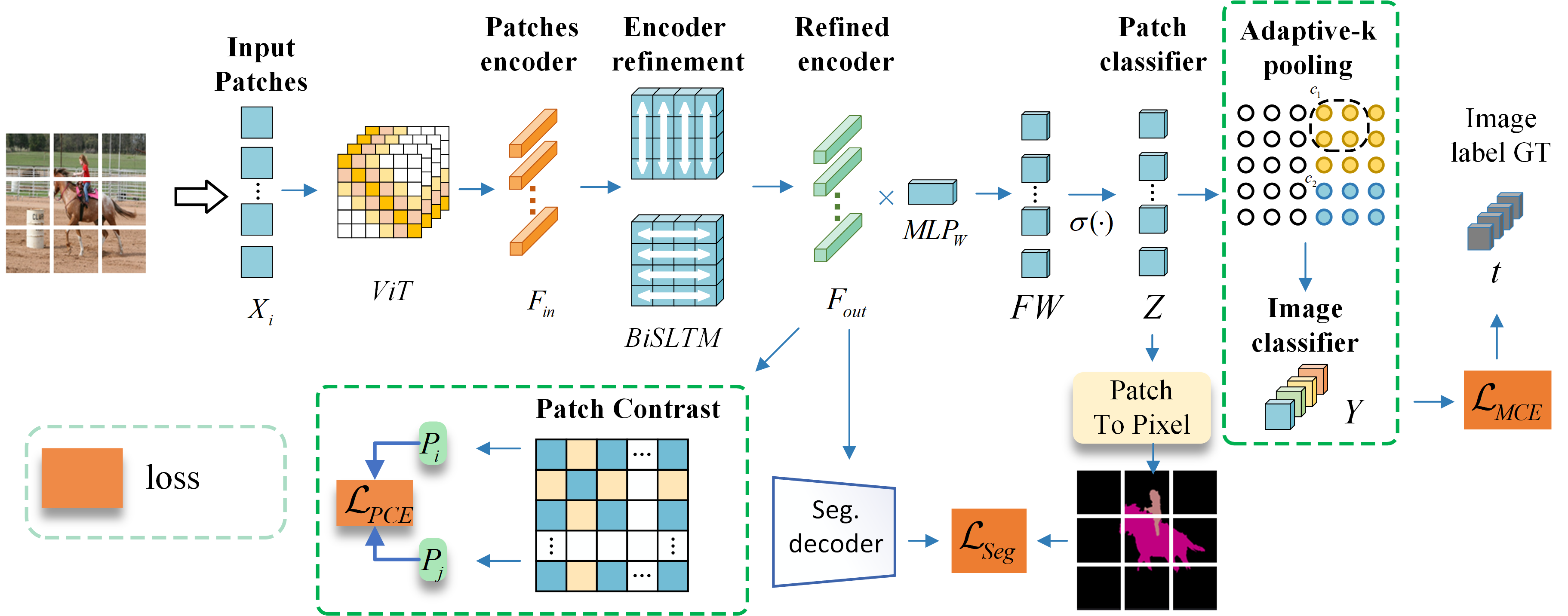}
\vspace{0.2cm}
   \caption{\tbl{APC infrastructure: ViT encodes patch embeddings, which are refined by BiLSTM. MLP and softmax produce patch-to-classifier predictions. The AKP module maps patch classifiers to an image classifier using image-level ground truth for supervision. After the Refined Encoder, the PCL module enhances patch similarity.}}
    \label{fig:TKP-PCL}
\end{center}

\end{figure*}
\subsection{Overall Framework} \label{sec:Overview}
Our proposed APC method mainly consists of three main components: the adaptive K pooling module, the image patch contrastive module, and the end-to-end encoder and decoder module, as illustrated in Figure~\ref{fig:framework}. We first compare with the previous without CAM multiple-stage framework, where our significant improvement lies in transforming the entire WSSS process into a single stage. By introducing the encoder and decoder modules, we enable the framework to directly predict segmentation results during training. Next, we delve into the main proper refinement component of APC. In the first stage, the input image is divided into fixed-size patches, and then a ViT encoder is used to generate an embedding for each patch. Each patch here is represented as an embedding space vector. To further enhance the performance of the patch classifier, adaptive K pooling is employed for patch selection specific to certain categories, and these selected patches are then used for prediction. Additionally, we introduce the patch contrastive module, which adjusts the cosine similarity of patch embeddings belonging to the same categories to be closer and that of patch embeddings belonging to different categories to be farther apart. Through the aforementioned modules, our method obtains prediction results, which are directly used as final labels. \tbl{After refinement with the Conditional Random Field (CRF) \cite{krahenbuhl2011efficient}, 
which is trained on PascalVOC in a fully supervised manner and introduced to WSSS by \cite{kolesnikov2016seed}, the results are directly used as the final output for verification, eliminating the need for an additional DeepLab stage.}

\subsection{Model Structure} \label{sec:model}

This section provides a comprehensive explanation regarding the generation of patch embeddings and the design of the end-to-end single-stage WSSS task. The illustration of details about this method is shown in Figure~\ref{fig:TKP-PCL}. The input image $X_{in}\in \mathbb{R}^{h\times w \times 3}$ is firstly divided into $s$ fixed-size input patches $X_{patch}\in \mathbb{R}^{d\times d \times 3}$, where $s = \frac{hw}{d^{2}}$. Then, the input patches $X_{patch}$ are passed into ViT-based encoder to generate the patch embeddings $F_{in}\in \mathbb{R}^{s\times e}$. Next, HV-BiLSTM is used to enhance the representation of $F_{in}$ and output refined patch embeddings $F_{out}$ with the same size as $F_{in}$. Given the refined patch embedding $F_{out}$, we use a weight $W\in \mathbb{R}^{e\times\mathcal{|C|}}$ and a softmax activation function to predict the class $c\in\mathcal{C}$ of each patch ~\tbl{in Eq.~\ref{eq:1}}, where $|\mathcal{C}|$ is the total number of classes:

\begin{equation}
\begin{aligned}
    Z=\text{softmax}(F_{out}W).
\end{aligned}
\label{eq:1}
\end{equation}
The output $Z\in \mathbb{R}^{s\times|\mathcal{C}|}$ represents the prediction scores of each class for each patch. To robustly project from patch-level predictions to image-level predictions, we propose an adaptive K pooling module, described in Sec.\ref{sec:top-K_Pooling}. The adaptive K module selects the top  $k$ patches $\{\bar{Z}^{c}_{i=1:k}\}$ with the highest value in each category and computes the average value as the prediction score for image-level classification~\tbl{in Eq.~\ref{eq:2}}:

\begin{equation}
\begin{aligned}
    y_c = \frac{1}{k} \sum_{i=1}^{k}\bar{Z}^{c}_{i},
\end{aligned}
\vspace{1em}
\label{eq:2}
\end{equation}

where $y_c$ is the projected image-level prediction score of class $c$. Adaptive K pooling ensures the final image prediction results are not dominated by any misclassified patches, thereby further improving the mapping from patch-level prediction to image-level prediction. We compute the distance between each patch pair using cosine similarity on the patch embeddings after refinement, and then apply contrastive learning, detailed in Section \ref{sec:PatchContrast}. Simultaneously, we introduce a decoder head to merge multi-level feature patch embeddings for prediction, implemented through a simple MLP layer. The decoder's prediction is combined with our patch prediction category result to calculate the segmentation loss \(L_{\text{seg}}\), where the predicted category of each patch is directly mapped onto each pixel within the patch as the pixel's category for calculating the segmentation loss. For the segmentation loss \(L_{\text{seg}}\), we utilize the commonly used cross-entropy loss. As illustrated in Figure~\ref{fig:TKP-PCL}, the supervision for the segmentation branch is the label refined through the Patch to Pixel module.

Finally, we minimize the error between the predicted image labels $y_{c}$ and ground-truth labels $t_{c}$ by using the multi-label classification prediction error (MCE)~\tbl{in Eq.~\ref{eq:3}}:
\begin{equation} 
\begin{aligned}
\mathcal{L}_{MCE}&=\frac{1}{|\mathcal{C}|}\sum_{c\in\mathcal{C}}{BCE(t_c,y_c)}\\
&=-\frac{1}{\mathcal{|C|}}\sum_{c\in\mathcal{C}}{t_c\log(y_c)+(1-t_c)\log(1-y_c)}.
\end{aligned}
\vspace{1em}
\label{eq:3}
\end{equation}
We use binary labels and $C$ independent Bernoulli distributions to model multi-label classification. With \td{$C$ representing the number of classes}, $C$ binary cross-entropy losses (BCE) measure the dissimilarity between predicted and ground-truth labels.

\subsection{Patch Embeddings with Adaptive K Pooling} \label{sec:top-K_Pooling}
The motivation behind using Adaptive K pooling is to facilitate the mapping between patch-level classification and image-level classification. In previous work~\cite{rossetti2022max}, Global Max Pooling (GMP) only selects the patches with the highest prediction scores for each class, which may result in inaccurate mapping of patch-level predictions to image-level classification. In our experiments, we observed that some patches are misclassified with high prediction scores, and when there are multiple objects of the same categories in an image, a single patch as a prediction score may not effectively represent the overall mapping of the image to these categories. Therefore, to achieve a more robust mapping between patch-level and image-level classification, we average the prediction scores of the top $k$ patches in each category as the prediction score for image-level classification. Additionally, we propose an adaptive algorithm to automatically select the value of $k$ based on different scenarios, instead of using a fixed value of $k$. This allows for more flexible adaptation to various scenarios.

The illustration of details about this method is shown in Figure~\ref{fig:TKP-PCL}. The output $Z\in \mathbb{R}^{s\times|\mathcal{C}|}$ represents the prediction scores for each class for each patch. We replace the original max pooling layer with an adaptive K pooling layer to map the patch category predictions of the entire image to image category predictions. Regarding the selection of K, we propose an adaptive K pooling module, described in Algorithm~\ref{alg:AK}. \tbl{We use an adaptive algorithm to select $k$ patches as candidates for pooling. Specifically, we first sort the patch prediction scores for a given category $c_i$ from high to low. Then, we select patches with sufficiently high scores as candidates to ensure that we can optimize not only the region corresponding to the patch with the highest score but also multiple regions with high scores.}

\begin{algorithm}
\caption{{\small{Adaptive K selection}}}\label{alg:AK}
\KwIn{$input\_tensor$ of patch-to-categories prediction scores, number of categories $C$}
\KwOut{$AdaptiveK\_candidate$ after adaptive K selection}
\BlankLine
$AdaptiveK\_candidate \gets \emptyset$\\
\BlankLine
\ForEach{category $c_i$ in $C$}{
    $selected\_elements \gets$ top-1 values of $input\_tensor$ for category $c_i$\\
    \BlankLine
    \For{$i \gets 2$ \KwTo $K$}{
        $current\_elements \gets$ top-$i$ values of $input\_tensor$ for category $c_i$\\
        $mean\_current \gets$ compute mean of $current\_elements$\\
        $mean\_selected \gets$ compute mean of $selected\_elements$\\
        \If{$\frac{mean\_current}{mean\_selected} > \theta$}{
            $selected\_elements \gets current\_elements$\\
        }
    }
    $AdaptiveK\_candidate[c_i] \gets selected\_elements$\\
}
\Return $AdaptiveK\_candidate$
\end{algorithm}
In our observation, we found that for each patch generating predictions for a specific category $Z$, in images containing multiple objects, besides the patch with the maximum prediction score, patches with prediction scores close to the maximum score also represent the category well. We designed an adaptive algorithm that automatically expands the selected set when it detects patches with prediction scores close to the maximum score and the difference between them is less than \(\theta\).

\subsection{Patch Contrastive Learning} \label{sec:PatchContrast}

To enhance the representation of patch embeddings and thus improve the accuracy of prediction, the patch contrastive learning (PCL) is used to narrow the distance between patch embeddings with high prediction scores in a specific category $c$ and also expand the distance between patch embeddings with high scores and the low confidence patch embeddings in the same category. In this work, cosine similarity is used to measure the distance between patch embeddings~\tbl{in Eq.~\ref{eq:cos}}:

\begin{equation} 
\begin{aligned}\label{eq:cos}
S(F_{out}^{i}, F_{out}^{j}) = \frac{F_{out}^{i} \cdot F_{out}^{j}}{\|F_{out}^{i}\|\|F_{out}^{j}\|},
\end{aligned}
\vspace{1em}
\end{equation}
where a higher similarity value means two patch embeddings are closer to each other, and a lower value indicates a further distance. To represent the similarity more explicitly, \tbl{as shown in Eq.~\ref{eq:5},} we normalize the range of value between 0 and 1 via:

\begin{equation}
\begin{aligned}
    \bar{S}(F_{out}^{i}, F_{out}^{j}) = \frac{1 + {S}(F_{out}^{i}, F_{out}^{j})}{2}.
\end{aligned}
\vspace{1em}
\label{eq:5}
\end{equation}
As illustrated in Figure~\ref{fig:TKP-PCL}, $F_{out}$ is used as the input of PCL module. Furthermore, the patch prediction score $Z_{i}^{c}$ and a threshold $\epsilon$ are used to determine whether a patch is a high confidence one $\mathcal{P}^{c}_{high} = \{F^{i}_{out}|Z_i^c > \epsilon\}$. Similarly, the patch with the lowest prediction score is considered as a low confidence one $\mathcal{P}^{c}_{low} = \{F^{i}_{out}|Z_i^c < (1 - \epsilon)\}$. Then, \tbl{in Eq.~\ref{eq:PCE},} the patch contrastive learning error (PCE) of a category $c$ can be expressed as:
 
\begin{equation} 
\begin{aligned}\label{eq:PCE}
\mathcal{L}_{PCE}^{c}&=\frac{1}{N_{pair}^{+}}\sum_{i=1}^{|\mathcal{P}^{c}_{high}|}\sum_{j=1,j\neq i}^{|\mathcal{P}^{c}_{high}|}(1-\bar{S}(F_{high}^{i}, F_{high}^{j}))\\
&+\frac{1}{N_{pair}^{-}}\sum_{m=1}^{|\mathcal{P}^{c}_{high}|}\sum_{n=1}^{|\mathcal{P}^{c}_{low}|} \bar{S}(F_{high}^{m}, F_{low}^{n}),
\end{aligned}
\vspace{1em}
\end{equation}
where $N_{pair}^{+}$ denotes the number of pairs of patches with high confidence, $N_{pair}^{-}$ denotes the number of pairs of high confidence and low confidence patches. $F_{high}$ and $F_{low}$ represent the feature embeddings of high confidence patch $\mathcal{P}^{c}_{high}$ and low confidence patch $\mathcal{P}^{c}_{low}$, respectively. 
\subsection{Overall loss} \label{sec:loss}

As illustrated in Figure~\ref{fig:TKP-PCL}, we have introduced multiple losses in the previous sections, including the contrastive learning \(L_{\text{pce}}\) loss, the classification \(L_{\text{mce}}\) loss, and the segmentation loss \(L_{\text{seg}}\). The overall loss of our APC\tbl{, as shown in Eq.~\ref{eq_loss},} is the weighted sum of \(L_{\text{pce}}\), \(L_{\text{mce}}\), and \(L_{\text{seg}}\): 
\begin{equation}
    \label{eq_loss}
    \mathcal{L} = \mathcal{L}_{mce} + \lambda_{1}\mathcal{L}_{seg} + \lambda_{2}\sum_{c\in\mathcal{C}}\mathcal{L}_{PCE}^{c}, 
\end{equation}
where $\lambda_1$, and $\lambda_2$ balance the contributions of different losses.

\section{Experiments}\label{sec4}
\label{sec:Experiments}

In this section, we describe the experimental settings, including dataset, evaluation metrics, and implementation details. We then compare our method with state-of-the-art approaches on PASCAL VOC 2012 dataset~\cite{everingham2010pascal} and MS COCO 2014~\cite{lin2014microsoft}. Finally, ablation studies are performed to validate the effectiveness of crucial components in our proposed method.


\begin{table*}[!h]

  \centering
  
  \caption{ \textbf{Impact of Image Size in the Inference Stage}: \tbl{Performance is evaluated on the VOC \texttt{val} set. The default settings are highlighted in gray as the baseline.}}
  \label{tab:pa}
  \begin{subtable}{0.7\textwidth}
    \setlength{\tabcolsep}{4mm}
    \centering
\begin{tabular}{l|ccc}
  \toprule
  Size     & $Seg.$ & Inference Time (seconds) & Memory (GB) \\ \midrule
  640$^2$          & 71.6   & 369   & 11.1     \\
  800$^2$           & 72.1   & 426   & 11.9    \\
  \rowcolor[HTML]{eaeaea}
  960$^2$           & 72.3   & 483   & 13.7    \\
  1120$^2$         & 72.0   & 920   & 18.8    \\ \bottomrule
\end{tabular}

  \end{subtable}

\end{table*}

\begin{figure}[h]
\centering
\includegraphics[width=1\linewidth]{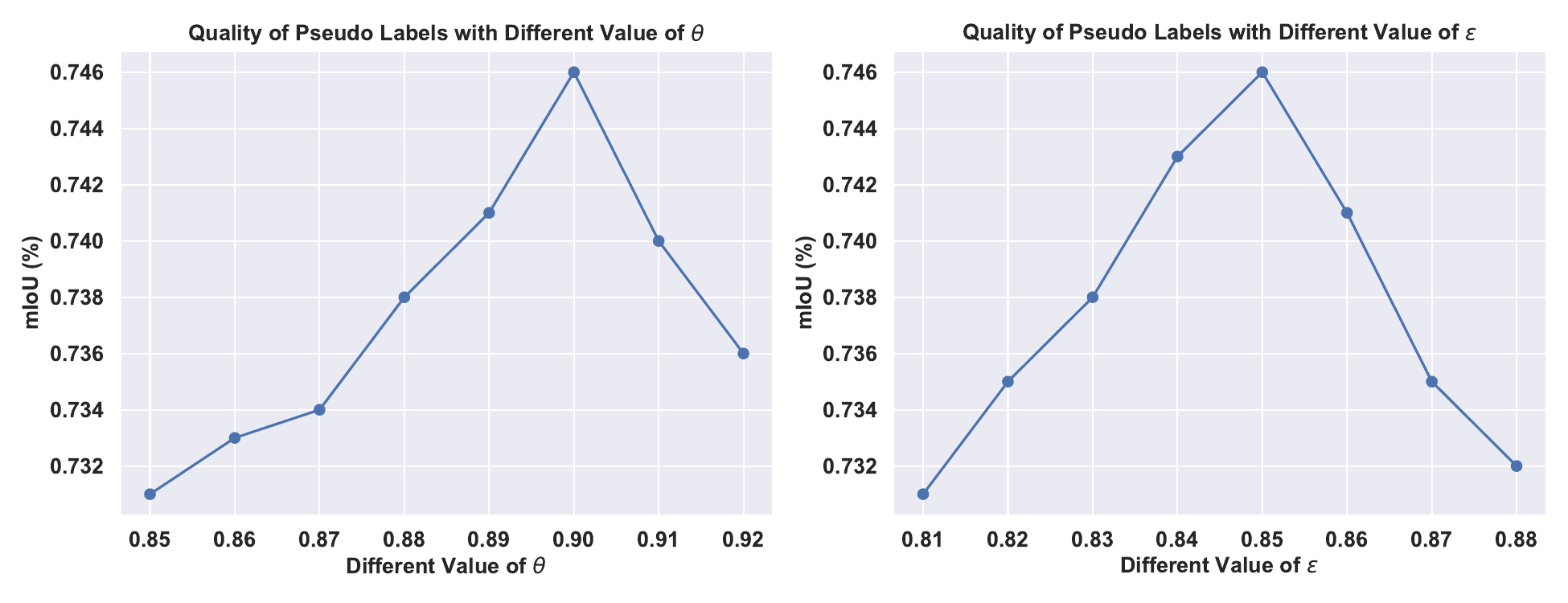}
\vspace{-0.6cm} 
\caption{The performance comparison of selecting different values of $\theta$ and $\epsilon$.}
\label{fig:diff}
\end{figure}

\begin{figure}[h]
\centering
\includegraphics[width=1\linewidth]{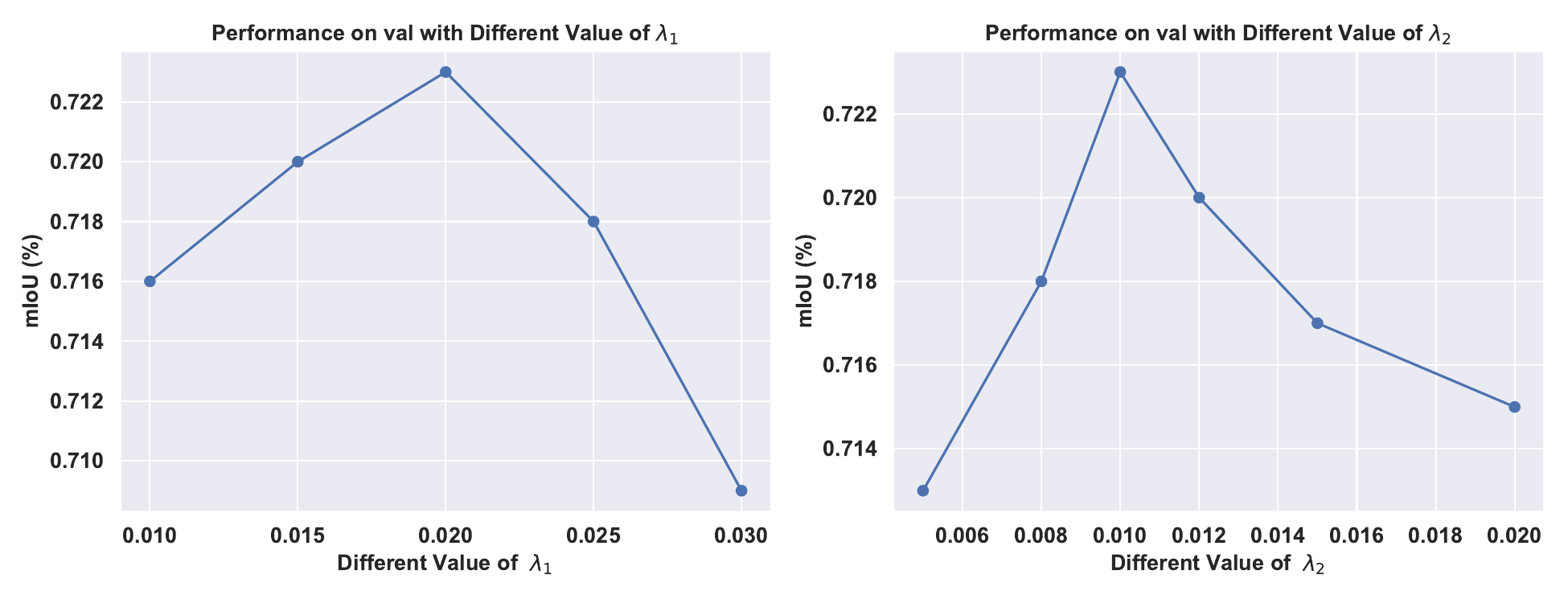}
\vspace{-0.6cm} 
\caption{The performance comparison of selecting different values of $\lambda_1$ and $\lambda_2$.}
\label{fig:diff_la}
\end{figure}

\begin{table}[!ht]
\centering
\caption{Results of predicted pseudo masks on PASCAL VOC 2012 train.}
\label{tab:vocbpm}

\begin{tabular}{@{}ccccc@{}}
\toprule
Method & Pub. & Backbone & mIoU (\%) \\
\midrule
SEAM~\cite{wang2020self} & CVPR20 & V1-RN38 & 63.6 \\
CONTA~\cite{zhang2020causal} & NeurIPS20 &V1-RN38 & 67.9 \\
CDA~\cite{su2021context} &ICCV21& V1-RN38 &66.4 \\
EPS \cite{lee2021railroad}&CVPR21&V2-RN101&71.4\\
Yao \emph{et al.} \cite{yao2021non}&CVPR21&V2-RN101&68.3\\
AuxSegNet \cite{xu2021leveraging}&ICCV21&V1-RN38&69.0\\
AdvCAM~\cite{lee2022anti} &PAMI22& V2-RN101&69.9\\
PPC~\cite{du2022weakly} & CVPR22 & ResNet38 & 61.5  \\
SIPE~\cite{chen2022self} & CVPR22 & ResNet50 & 58.6  \\
AFA~\cite{ru2022learning} & CVPR22 & MiT-B1 & 66.0 \\
ViT-PCM~\cite{dosovitskiy2020image} & ECCV22 & ViT-B/16 & 71.4 \\
ToCo~\cite{ru2023token} & CVPR23 & ViT-B/16 & 72.2 \\
USAGE~\cite{Peng_2023_ICCV} & ICCV23 & ResNet38 & 72.8\\
FPR~\cite{chen2023fpr} & ICCV23 & ResNet38 & 68.5\\

\midrule
\textbf{APC (Ours)} &  & ViT-B/16 & \textbf{74.6}  \\
\bottomrule
\end{tabular}

\end{table}

\subsection{Experimental Settings}

\textbf{Dataset and Evaluated Metric.} Our experiments are conducted on two benchmark datasets: PASCAL VOC 2012~\cite{everingham2010pascal}, which includes $21$ classes, and MS COCO 2014~\cite{lin2014microsoft}, featuring $81$ classes. Each dataset also contains an extra class for background. The PASCAL VOC 2012 dataset is commonly expanded using the SBD dataset~\cite{hariharan2011semantic}. For training on PASCAL VOC 2012, we use 10,582 images with image-level labels and 1,449 images for the validation set. In the case of the MS COCO 2014 dataset, approximately 82k images are used for training and around 40k images for validation, with the training images having only image-level annotations. The performance is evaluated using the mean Intersection-Over-Union (mIoU) metric. Our APC approach demonstrates substantial improvements in segmentation results on both PASCAL VOC 2012 and MS COCO 2014 datasets.

\textbf{Implementation Details.} In our experiments, we utilize the ViT-B/16 model as the backbone for the ViT encoder. During the training process, input images are initially resized to 384×384, as suggested in Kolesnikov et al. (2016)~\cite{kolesnikov2016seed}, and subsequently partitioned into 24×24 small patches for the ViT encoder. For the inference stage, we resize the image to 960×960 and \tbl{the impact of image size is assessed on the VOC val set, with default settings indicated in baseline color gray, as illustrated in Table~\ref{tab:pa}. We observe that the increase in memory usage from $640$ to $800$ is relatively small, rising from 11.1 GB to 11.9 GB. This may be attributed to the intermediate layers not reaching their memory bottleneck and the GPU’s effective memory compression or optimization. However, when increasing from $960$ to $1120$, memory usage jumps from 13.7 GB to 18.8 GB, accompanied by a drop in performance. Building on insights from prior work~\cite{kabir2023machine,bolon2020feature}, we learn that higher data complexity does not necessarily lead to better results. In fact, reducing the complexity from $1120$ to $960$ not only decreases data complexity but also improves overall performance. The model is trained with a batch size of 16 and for a maximum of 15 epochs, utilizing two NVIDIA 4090 GPUs.} We employ the Adam optimizer and schedule the learning rate as follows: the learning rate is set to $10^{-3}$ for the first two epochs, and then to $10^{-4}$ for the remaining epochs. We set the threshold $\epsilon$ to 0.85 to determine high-confidence patches in PCE, and $k$ is set to 6 for adaptive-K pooling. \tbl{The default weight coefficients $\lambda_1$ and $\lambda_2$ are set to 0.02 and 0.01, respectively, as shown in Figure~\ref{fig:diff_la}. We employed grid search to determine the optimal values for $\lambda_1$ and $\lambda_2$ within specified ranges. Initially, we removed the $\lambda_2$ module and enumerated possible values for $\lambda_1$ within [0, 1], adjusting the step size from larger to smaller increments until we identified the optimal value for $\lambda_1$. We then followed a similar approach for $\lambda_2$. This stepwise adjustment aimed to isolate the effect of each parameter on model performance. Finally, we applied the learned values of $\lambda_1$ and $\lambda_2$ to the MS COCO dataset, where we observed SOTA results as well. And $\theta$ is set to 0.9. As shown in Figure~\ref{fig:diff}, the performance is compared by selecting different values for $\theta$ and $\epsilon$. During the validation phase, we do not require additional DeepLab training for full supervision. Instead, we directly use the model's prediction results as the final labels and validate them on the validation set.}

\begin{table}[ht]
\centering
\caption{Final Semantic segmentation performance for training on Pascal VOC 2012 val.}
\label{tab:vocseg}

\begin{tabular}{@{}ccccc@{}}

\toprule
Model & Pub. & Backbone & mIoU (\%)\\
\midrule
\textbf{\large \textit{Multiple-Stage method}}
\\

MCTformer~\cite{xu2022multi} & CVPR22 & DeiT-S & 61.7 \\
SIPE~\cite{chen2022self} & CVPR22 & ResNet50 & 58.6\\
ViT-PCM~\cite{dosovitskiy2020image} & ECCV22 & ViT-B/16 & 69.3 \\
USAGE~\cite{Peng_2023_ICCV} & ICCV23 &ResNet38 & 71.9\\
SAS~\cite{kim2023semantic} & AAAI23 & ViT-B/16& 69.5\\
FPR~\cite{chen2023fpr} & ICCV23 & ResNet38 & 70.0\\
\\
\textbf{\large \textit{End-To-End Single-Stage}}
\\
AFA~\cite{ru2022learning} & CVPR22 & MiT-B1 & 63.8 \\
ToCo~\cite{ru2023token} & CVPR23 & ViT-B/16 & 70.5 \\
TSCD~\cite{Xu_Wang_Sun_Xu_Meng_Zhang_2023} & AAAI23 & MiT-B1& 67.3\\
\midrule
\textbf{APC (End-To-End) } &  & ViT-B/16 & \textbf{72.3 } \\
\bottomrule
\end{tabular}

\end{table}

\subsection{Comparisons with State-of-the-art}
\textbf{Comparison of Pseudo Masks.} We evaluated the performance of our APC method on the PASCAL VOC 2012 dataset for pseudo mask prediction, using 10,582 images with only image-level labels to train the segmentation network. Subsequently, we conducted segmentation predictions on these 10,582 images. As demonstrated in Table~\ref{tab:vocbpm}, our APC method outperforms other state-of-the-art techniques, encompassing both multi-stage and single-stage approaches, by achieving an mIoU value of 74.6\% on the training data. This achievement is credited to our proposed adaptive-K pooling method, which dynamically selects an optimal number of K patches for category prediction scores, effectively addressing the challenge of mapping patch-level classification to image-level classification using a single patch's prediction score. Additionally, the incorporation of patch contrastive learning further boosts the overall performance.

\begin{table}[ht]
\centering
\caption{ Final Semantic segmentation performance for training on MS COCO 2014 val set.}
\label{tab:cocoseg}

\begin{tabular}{@{}ccccc@{}}

\toprule
Model & Pub. & Backbone & mIoU (\%)\\
\midrule
\textbf{\large \textit{Multiple-Stage method}}
\\

MCTformer~\cite{xu2022multi} & CVPR22 & Resnet38 & 42.0 \\
ViT-PCM~\cite{dosovitskiy2020image} & ECCV22 & ViT-B/16 & 45.0 \\
SIPE~\cite{chen2022self} & CVPR22 & Resnet38 & 43.6\\
OCR~\cite{cheng2023out} & CVPR23 & ViT-B/16& 42.5\\
SAS~\cite{kim2023semantic} & AAAI23 & ViT-B/16& 44.8\\
FPR~\cite{chen2023fpr} & ICCV23 & ResNet38 & 43.9\\

\\
\textbf{\large \textit{End-To-End Single-Stage}}
\\
AFA~\cite{ru2022learning} & CVPR22 & MiT-B1 & 38.9 \\
ToCo~\cite{ru2023token} & CVPR23 & ViT-B/16 & 42.3 \\
TSCD~\cite{Xu_Wang_Sun_Xu_Meng_Zhang_2023} & AAAI23 & MiT-B1& 40.1\\

\midrule
\textbf{APC (Ours)} &  & ViT-B/16 & \textbf{45.7 } \\
\bottomrule
\end{tabular}

\end{table}

\textbf{Comparison of Segmentation Results.} In this study, we compared the final segmentation performance of our proposed APC end-to-end method on the PASCAL VOC 2012 and MS COCO 2014 datasets. As shown in Table~\ref{tab:vocseg}, consistent with other comparative methods, we utilized 10,582 images with only image-level labels as the training set to train our APC model. For validation, we employed 1,449 images to assess our final semantic segmentation performance, where our end-to-end single-stage method outperformed both recent end-to-end approaches and multiple-stage methods. Notably, the end-to-end method significantly surpasses multiple-stage methods in computational efficiency, offering substantial savings in capacity. These details will be further elaborated in Section~\ref{sec:as}. Regarding the MS COCO 2014 dataset, we adopted a similar approach, using approximately 82k images for training and around 40k images for validation to evaluate the final semantic segmentation performance. As indicated in Table~\ref{tab:cocoseg}, our method continued to exhibit superior performance.

\subsection{Ablation Studies}
\label{sec:as}

\begin{table}[h!]
\centering
\caption{\tbl{Ablation studies on computational efficiency and final performance for Pascal VOC 2012 and MS COCO 2014 datasets.}}
\label{tab:abt}

\begin{tabular}{cccccc}
\toprule
Dataset & Baseline (ViT-B) & AKP & PCL & Time (Hours) & mIoU (\%) \\
\midrule
VOC & \checkmark &  &  & 4\(\pm\)1 & 68.2  \\
VOC & \checkmark & \checkmark &  & 3\(\pm\)0.5 & 70.3  \\
VOC & \checkmark & \checkmark & \checkmark & 1.5\(\pm\)0.5 & 72.3  \\
\midrule
COCO & \checkmark &  &  & 9\(\pm\)2 & 42.8  \\
COCO & \checkmark & \checkmark &  & 7\(\pm\)1 & 44.2  \\
COCO & \checkmark & \checkmark & \checkmark & 3\(\pm\)1 & 45.7  \\
\bottomrule
\end{tabular}
\end{table}

\begin{table}[h!]
\centering
\caption{\tbl{Efficiency performance of APC compared to others on Pascal VOC 2012 validation set.}}
\label{tab:e2etime}

\vspace{-0.5em}
\begin{tabular}{l|ccc|c}
\toprule
\textbf{Method} & \multicolumn{3}{c|}{Time (Hours)} & \textbf{Val (\%)}  \\ 
\midrule
\multicolumn{5}{l}{\textbf{Multiple-Stage Methods}} \\ 
\midrule
ViT-PCM~\cite{rossetti2022max} & \multicolumn{3}{c|}{8\(\pm\)1} & 69.3\\ 
\midrule
\multicolumn{5}{l}{\textbf{End-to-End Methods}} \\ 
\midrule
AFA~\cite{ru2022learning} & \multicolumn{3}{c|}{2.5\(\pm\)0.5} & 63.8 \\
ToCo~\cite{ru2023token} & \multicolumn{3}{c|}{2\(\pm\)0.5} & 70.5 \\

\rowcolor[HTML]{EFEFEF} 
\textbf{APC (Ours)} & \multicolumn{3}{c|}{\textbf{1.5\(\pm\)0.5}} & \textbf{72.3} \\ 
\bottomrule
\end{tabular}

\vspace{-1.5em}
\end{table}

\textbf{Performance and Computational Efficiency.} Table~\ref{tab:abt} demonstrates the superiority of our APC in terms of computational performance. \tbl{We conducted ablation experiments comparing the baseline ViT-based end-to-end single-stage module (without AKP and PCL), the Adaptive-K Pooling (AKP) module, and the Patch Contrastive Learning (PCL) module. Baed on findings from prior research~\cite{ru2022learning,ru2023token}, traditional multi-stage methods are less efficient than end-to-end approaches. This is because multi-stage methods typically involve training pseudo-labels in the first stage, followed by their use as training data for fully supervised Deeplab network training, which consumes significant computational resources. Therefore, we also include a comparison with other end-to-end methods in Table}~\ref{tab:e2etime} \tbl{to evaluate the runtime performance of our approach. To ensure fairness in experimentation, all comparisons were conducted on a single NVIDIA 4090 GPU machine.}
As shown in Table~\ref{tab:abt}, \tbl{In Pascal VOC 2012, our baseline ViT-based end-to-end single-stage module required 4 hours to complete. Incorporating the Adaptive-K Pooling (AKP) module further reduced the runtime to approximately 3 hours. Finally, with the addition of contrastive learning, significant performance improvements were observed, with the model typically converging within 15 epochs. The total training time was reduced to approximately 1.5 hours. The same benefits are observed in alignment with the results on MS COCO 2014. The Experimental results demonstrate that our approach not only greatly enhances the final segmentation performance but also significantly improves computational efficiency, thereby conserving more computational resources.}

\begin{table}[ht]
\centering
\caption{\tbl{Ablation studies on different pooling strategies and keeping other components consistent for final semantic segmentation performance on Pascal VOC 2012 val.}}
\label{tab:diff-pooling}

\begin{tabular}{ccccccc}
\toprule
Average pooling & Max-Pooling & Top-K  & AKP & mIoU(\%)\\
\midrule
\checkmark & &  &  & 61.7\% \\
& \checkmark &  &  & 70.5\% \\
&  & \checkmark & & 71.8\% \\
&  &  & \checkmark  & 72.3\% \\
\bottomrule
\end{tabular}

\end{table}

\begin{table}[h!]
\centering
\caption{Ablation studies on our APC model for the final semantic segmentation performance on the Pascal VOC 2012 val. The \textcolor{gray}{components} constitute the initial baseline.}
\label{tab:compab}

\begin{tabular}{ccccccc}
\toprule

\textcolor{gray}{Backbone (ViT)} &\textcolor{gray}{HV-BiLSTM} & \textcolor{gray}{CRF} & AKP & PCL  & Seg mIoU (\%)\\
\midrule
\checkmark & & &    & &62.6\% \\
\checkmark &\checkmark& &    & &65.3\% \\
\checkmark & \checkmark &\checkmark&    && 68.6\% \\
\checkmark &\checkmark& \checkmark & \checkmark & &70.3\% \\
\checkmark &\checkmark&  \checkmark&   \checkmark &\checkmark&72.3\% \\
\bottomrule
\end{tabular}

\end{table}

\textbf{The Impact of Different Pooling Strategies.} Table~\ref{tab:diff-pooling} presents ablation studies on various pooling strategies for final semantic segmentation performance. We compared the performance of average pooling layer, max pooling layer, fixed top-K pooling layer, and our proposed adaptive-K pooling layer. To ensure fair experimentation, we incorporated the PCL module and utilized an end-to-end framework in all cases. The only difference lies in the choice of pooling layer. The experiments were conducted on the Pascal VOC 2012 validation dataset. We observed that our proposed adaptive-K pooling achieved the highest performance.

\begin{figure*}[t]
\centering
\includegraphics[width=0.8\linewidth]{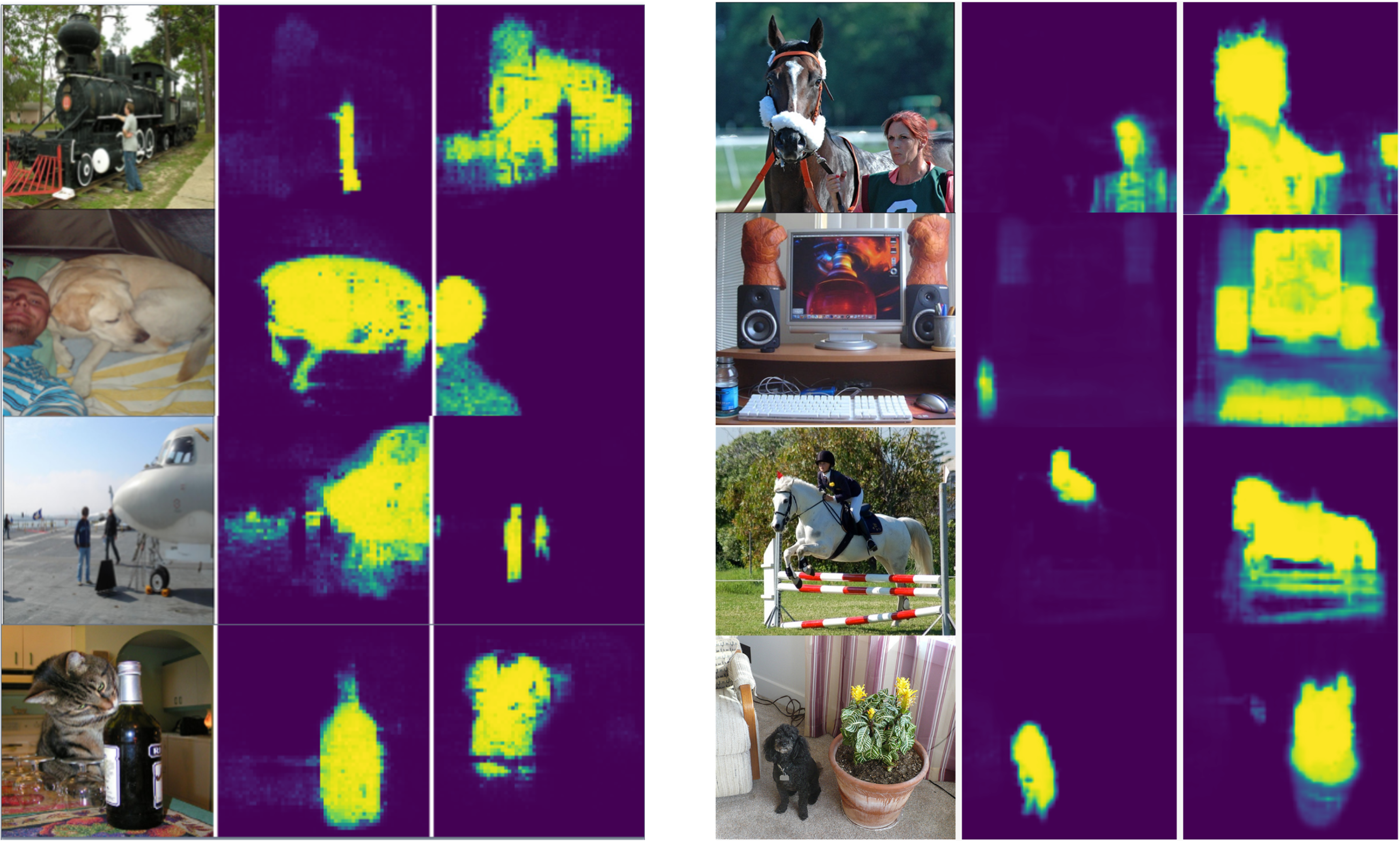}

\caption{\tbl{The sample illustrates probabilities highlighted by 60×60 heatmaps, which are generated from the patch to pixel module to produce pixel probabilities.}}
\label{fig:heatmap}

\end{figure*}

\begin{figure*}[t]
\centering
\includegraphics[width=\linewidth]{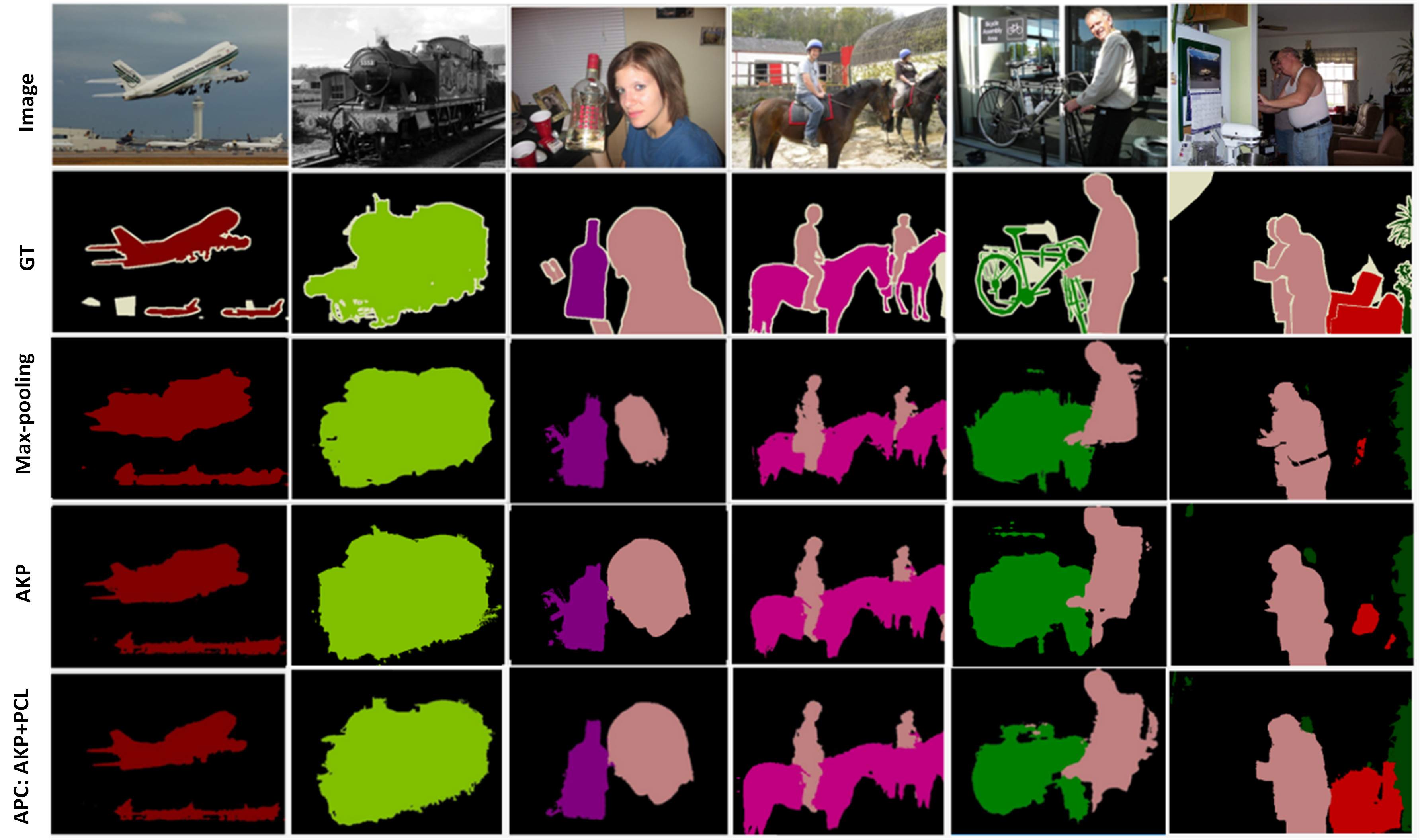}

\caption{The comparison of qualitative segmentation results from top to bottom: the images, GT, max-pooling (ViT-PCM~\cite{dosovitskiy2020image}), ViT+AKP, and our APC utilizing ViT+AKP+PCL.}
\label{fig:result}

\end{figure*}

\begin{table}[h]
\centering
\caption{Comparison of Inter-Class and Intra-Class Cosine Distances With and Without PCL. }
\label{tab:dis}
\begin{tabular}{lcc}
\toprule
\textbf{Method} & \textbf{Type}     & \textbf{Cosine Distance} \\
\midrule
With PCL        & Inter-class           & 0.9556   \\
With PCL        & Intra-class           & 0.5664  \\
Without PCL     & Inter-class           & 0.9473  \\
Without PCL     & Intra-class           & 0.7152  \\
\bottomrule
\end{tabular}
\label{tab:cos_distance}
\end{table}

\textbf{The Impact of Different component.} 
\tbl{The ablation studies of different components in our framework are presented in Table~\ref{tab:compab}. We designed the experiments to analyze the impact of HV-BiLSTM, CRF, AKP, and PCL. HV-BiLSTM and CRF serve as our baselines, marked in gray, as both are well-established methods from previous research with proven effectiveness for WSSS tasks.} Our key contributions, AKP and PCL, are  shown in Table~\ref{tab:compab}, AKP improves the final segmentation performance by 1.7\% mIoU compared to max pooling. \tbl{Additionally, by incorporating the proposed PCL, which applies contrastive learning at the patch level within each image, enhancing inter-class separability and improving intra-class consistency in the feature space, the segmentation task performance was further improved by 2.0\% mIoU. Meanwhile, we design a new metric to measure inter-class separability using cosine distance based on category embeddings. Specifically, we use patch embeddings from the intermediate results of our model, similar to the cosine distance in the PCL method. As shown in Table~\ref{tab:dis}, we observe that after applying PCL, the inter-class distance increases, while the intra-class distance decreases compared to before. This result verifies that our PCL module enhances inter-class separability and improves intra-class consistency.} Therefore, these two components significantly contribute to outperforming previous ViT-based methods in our approach.

\subsection{Visualization Results }
\textbf{Visualization of Heatmaps.} We present the results of heatmap visualization in Figure~\ref{fig:heatmap}, where probabilities are highlighted by 60×60 heatmaps. Pixels with values in yellow indicate the probability of belonging to the predicted class. \tbl{Through the heatmap results, we can see that in key areas of the object, the probabilities are high, while they gradually decrease at the edges. Our APC effectively segments the objects.}

\textbf{Segmentation Visualization Results.} We present the final semantic segmentation results in Figure~\ref{fig:result}. \tbl{Additionally, we compared these results with those of the previous max-pooling-based method, demonstrating a significant improvement in semantic segmentation performance. Specifically, we observe that in the results from max-pooling, the segmentation regions are highly concentrated. This is due to max-pooling being limited to the areas surrounding individual patches. For instance, in human segmentation, max-pooling tends to focus on regions around the face, such as the eyes and nose, while neglecting areas like the hair. With our AKP module, however, the segmentation of the person is significantly enhanced. Regarding the PCL module, we observe substantial improvements in object boundary segmentation. By leveraging inter-class and intra-class contrastive learning, PCL effectively enhances the segmentation of object edges.}

\section{Conclusion}\label{sec6}
In this work, we propose APC approach without using CAM for weakly supervised semantic segmentation. Unlike previous methods, APC utilizes adaptive-K pooling to select k patches for mapping patch-level classification to image-level classification, thus mitigating the issue of potential misclassified patches
. In addition, patch contrastive learning (PCL) is proposed to further enhance the feature embeddings of patches. In the same class, PCL aims to decrease the distance between patch embeddings with high confidence and increase the distance between embeddings with high confidence and embeddings with low confidence. By combining these two components, our method achieves state-of-the-art results in weakly supervised semantic segmentation tasks using only image-level labels.

\tbl{However, a limitation of our framework is that it relies solely on image information as input. In WSSS tasks, the category information for each image is typically available. If this information were incorporated as text and fused with image features through multimodal integration, it could potentially enhance segmentation performance, particularly in accurately identifying object boundaries. In comparison, our current approach depends exclusively on image data, which may not offer the most optimal solution.}

\tbl{In future work, we will further enhance segmentation performance by enriching the input information with semantic features and integrating category text features into the images for model training. However, the current network architecture presents challenges for directly utilizing text features in a multimodal fusion approach. Therefore, future studies should consider designing a network model that can effectively accommodate both text and image inputs.}

\bibliography{mybib}

\end{document}